\pdfoutput=1

\documentclass[11pt]{article}

\usepackage[final]{acl}

\usepackage{times}
\usepackage{latexsym}

\usepackage[T1]{fontenc}

\usepackage[utf8]{inputenc}

\usepackage{microtype}

\usepackage{inconsolata}

\usepackage[utf8]{inputenc} 
\usepackage[T1]{fontenc}    
\usepackage{hyperref}       
\usepackage{url}            
\usepackage{booktabs}       
\usepackage{amsfonts}       
\usepackage{nicefrac}       
\usepackage{microtype}      
\usepackage{xcolor}         
\usepackage{graphicx} 
\usepackage{subcaption}
\usepackage{tabularx}
\usepackage{multirow}
\usepackage{caption}
\usepackage{enumitem}
\usepackage{float}
\usepackage[linesnumbered,ruled,vlined]{algorithm2e}
\usepackage{amsmath}
\usepackage{array}
\usepackage{adjustbox}
\usepackage{caption}
\usepackage[framemethod=TikZ]{mdframed}
\usepackage{multicol}
\usepackage{framed}

\title{\textsc{SpreadsheetLLM}: Encoding Spreadsheets for Large Language Models}

\author{
Haoyu Dong\thanks{\ \ Corresponding author (hadong@microsoft.com).}\thanks{\ \ Equal contribution.}, 
Jianbo Zhao\footnotemark[2]\thanks{\ \ Work during internship at Microsoft.}, 
Yuzhang Tian\footnotemark[2]\footnotemark[3], 
Junyu Xiong\footnotemark[2]\footnotemark[3], 
Shiyu Xia\footnotemark[3], 
\\
\textbf{Mengyu Zhou,
Yun Lin\footnotemark[3],
José Cambronero,
Yeye He,
Shi Han,
Dongmei Zhang}
\\
Microsoft Corporation
}

\begin{document}
\maketitle
\begin{abstract}
Spreadsheets are characterized by their extensive two-dimensional grids, flexible layouts, and varied formatting options, which pose significant challenges for large language models (LLMs). In response, we introduce \textsc{SpreadsheetLLM}, pioneering an efficient encoding method designed to unleash and optimize LLMs’ powerful understanding and reasoning capability on spreadsheets. Initially, we propose a vanilla serialization approach that incorporates cell addresses, values, and formats. However, this approach was limited by LLMs' token constraints, making it impractical for most applications. To tackle this challenge, we develop \textsc{SheetCompressor}, an innovative encoding framework that compresses spreadsheets effectively for LLMs. It comprises three modules: structural-anchor-based compression, inverse index translation, and data-format-aware aggregation. It significantly improves performance in the spreadsheet table detection task, outperforming the vanilla approach by 25.6\% in GPT4’s in-context learning setting. Moreover, fine-tuned LLM with \textsc{SheetCompressor} has an average compression ratio of 25×, and achieves a state-of-the-art 78.9\% F1 score, surpassing the best existing models by 12.3\%. 
Finally, we propose Chain of Spreadsheet for downstream tasks of spreadsheet understanding and validate it in a new and demanding spreadsheet QA task. We methodically leverage the inherent layout and structure of spreadsheets, demonstrating that \textsc{SpreadsheetLLM} is highly effective across a variety of spreadsheet tasks.

\end{abstract}

\begin{figure}[htbp]
\label{overview}
    \includegraphics[width=0.45\textwidth]{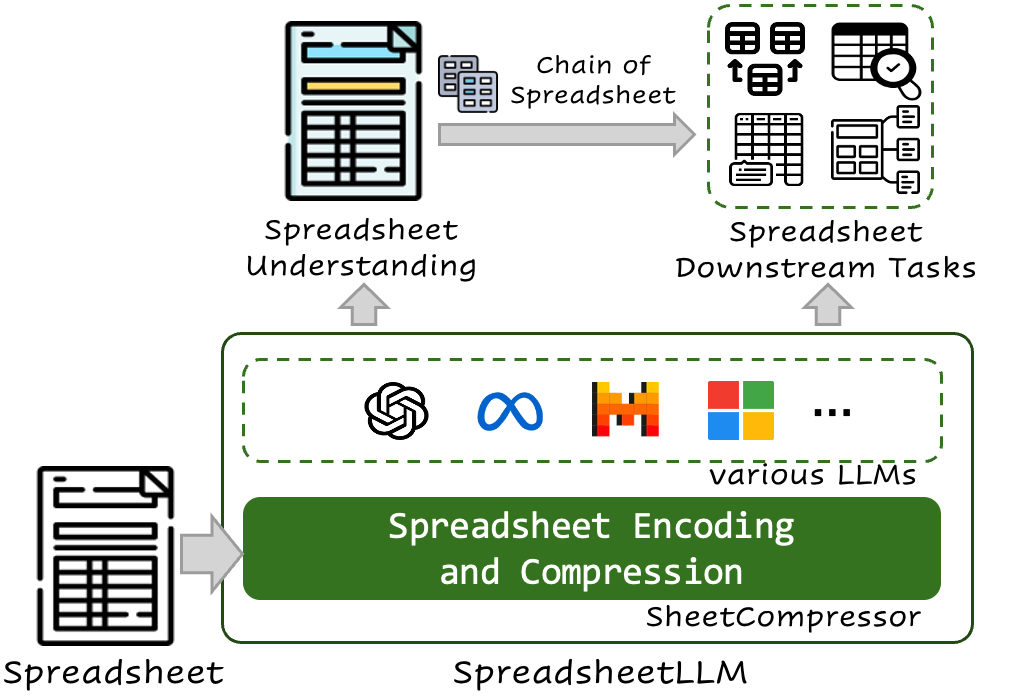}
    \caption{The \textsc{SpreadsheetLLM} pipeline.}   
    \vspace{-4mm}
\end{figure}

\vspace{-2mm}
\section{Introduction}
Spreadsheets are ubiquitous for data management and extensively utilized within platforms like Microsoft Excel and Google Sheets. Understanding spreadsheet layout and structure~\cite{dong2019tablesense,gol2019tabular,hulsebos2019sherlock,dou2018expandable,wang2021tuta,deng2022turl,chen2014integrating}, a longstanding challenge for traditional models, is crucial for effective data analysis and intelligent user interaction. Recently, the rapid development of Large Language Models (LLMs) has opened new frontiers in table processing~\cite{li2023table} and reasoning~\cite{cheng2022binding}. However, spreadsheets pose unique challenges for LLMs due to their expansive grids that usually exceed the token limitations of popular LLMs, as well as their inherent two-dimensional layouts and structures, which are poorly suited to linear and sequential input. Furthermore, LLMs often struggle with spreadsheet-specific features such as cell addresses and formats, complicating their ability to effectively parse and utilize spreadsheet data, as detailed in Appendix \ref{struggle}. 

\begin{figure*}[htbp]
    \centering
    \includegraphics[width=1\textwidth]{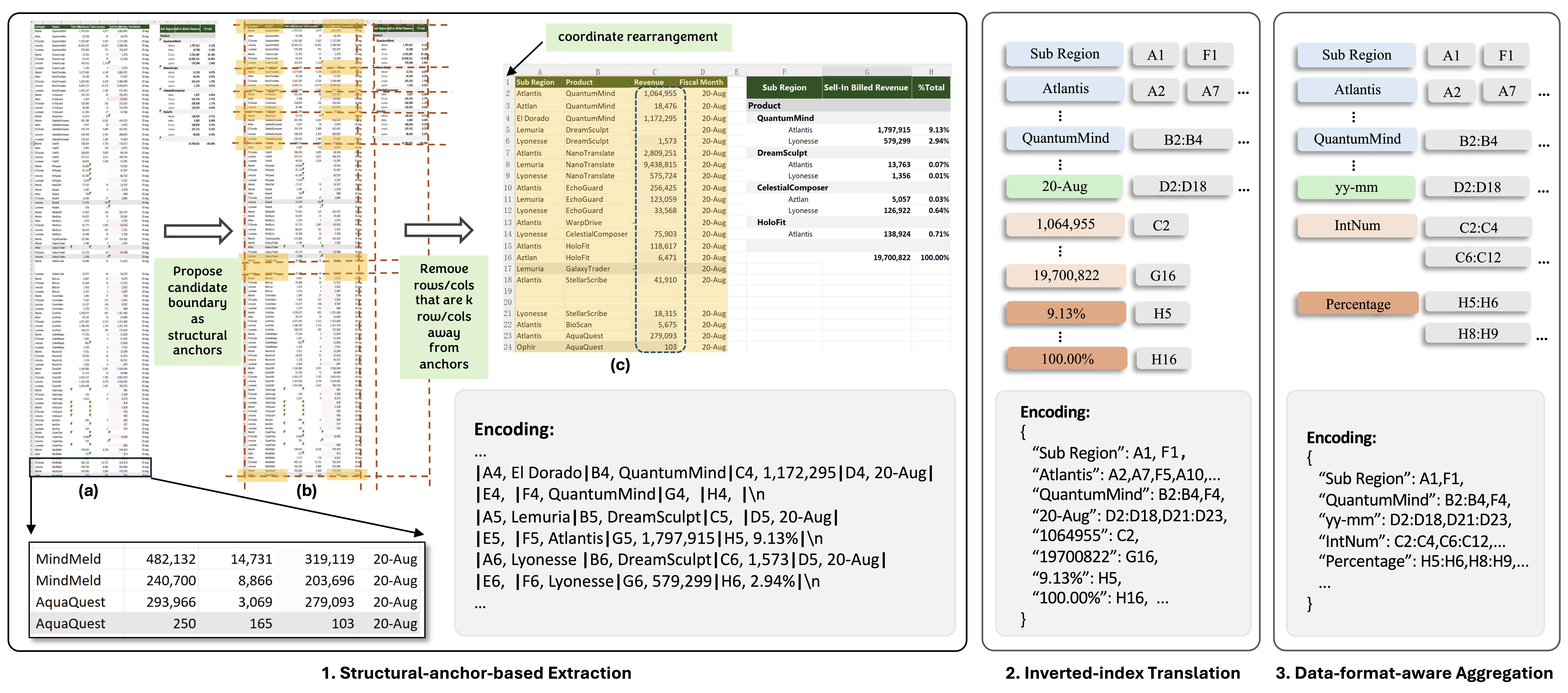}
    \caption{Illustration of the \textsc{SheetCompressor} framework. The original spreadsheet contains two tables, featuring numerous data entries or hierarchical headers. The completed spreadsheet consists of 576 rows and 23 columns, with a vanilla encoding of 61,240 tokens. Initially, we first extract cells using structural anchors, rearranging them into a smaller 24×8 sheet. Subsequently, we perform index-invert, removing empty cells. Finally, we aggregate cells based on data formats, achieving an extremely compact representation of the spreadsheet with only 708 tokens.} 
    \label{fig:method}
\end{figure*}

In this paper, we introduce \textsc{SpreadsheetLLM}, a pioneering framework to unleash and maximize the potential of LLMs for spreadsheet understanding and reasoning.
We initially propose a vanilla encoding method to serialize spreadsheets into sequences, augmenting the Markdown encoding method by including essential cell addresses and (optional) formats. 
Furthermore, large spreadsheets that exceed the token limits of LLMs not only limit their processing but also, as observed in prior studies, degrade accuracy performance as the size increases~\cite{liu2024lost}. To address this challenge, we propose \textsc{SheetCompressor}, featuring a novel encoding framework comprising three portable modules:

\textbf{1) Structural Anchors for Efficient Layout Understanding:} Observations indicate that large spreadsheets often contain numerous homogeneous rows or columns, which contribute minimally to understanding the layout and structure (see left panel in Figure~\ref{fig:method} (a)). To address this, we identify structural anchors—heterogeneous rows and columns at possible table boundaries that offer substantial layout insights, as depicted in Figure~\ref{fig:method} (b). Then we remove distant, homogeneous rows and columns, producing a condensed "skeleton" version of the spreadsheet, as illustrated in Figure~\ref{fig:method} (c).

\textbf{2) Inverted-Index Translation for Token Efficiency:} The vanilla encoding method becomes token-consuming when handling spreadsheets with numerous empty cells and repetitive values, as shown in Figure~\ref{fig:method} (c). To improve efficiency, we depart from traditional row-by-row and column-by-column serialization and employ a lossless inverted-index translation in JSON format. This method creates a dictionary that indexes non-empty cell texts and merges addresses with identical text, optimizing token usage while preserving data integrity.

\textbf{3) Data Format Aggregation for Numerical Cells:} Adjacent numerical cells often share similar number formats. Recognizing that exact numerical values are less crucial for grasping spreadsheet structure, we extract number format strings and data types from these cells. Then adjacent cells with the same formats or types are clustered together. This method is visualized in the right example of Figure~\ref{fig:method}, where rectangular regions are represented by uniform format strings and data types, streamlining the understanding of numerical data distribution without excessive token expenditure.

We conducted a comprehensive evaluation of our method on a variety of LLMs. Our experiments show that \textsc{SheetCompressor} significantly reduces token usage for spreadsheet encoding by 96\%. Moreover, \textsc{SpreadsheetLLM} has shown exceptional performance in spreadsheet table detection, the foundational task of spreadsheet understanding, surpassing the previous SOTA method by 12.3\%~\cite{dong2019tablesense}. We also applied \textsc{SpreadsheetLLM} to a representative spreadsheet QA task. Inspired by the Chain of Thought (CoT) methodology~\cite{zheng-etal-2023-chain, jiang2023structgpt}, we propose Chain of Spreadsheet (CoS) to decompose spreadsheet reasoning into a table detection-match-reasoning pipeline. It significantly outperformed existing SOTA methods for table QA~\cite{liu2022tapextablepretraininglearning, cheng2022binding}.
Our primary contributions are summarized as follows:
\vspace{-2mm}
\begin{itemize}
\item We propose \textsc{SpreadsheetLLM}, the first work that substantially leverage
LLMs for understanding and analyzing spreadsheet data. To address challenges in scale, diversity, and complexity of spreadsheets, we propose \textsc{SheetCompressor}, an innovative encoding framework to compress spreadsheets for LLMs with efficient encoding. 
\vspace{-2mm}
\item We fine-tune a variety of cutting-edge LLMs to achieve optimal performance on spreadsheet table detection and demonstrate the high effectiveness of \textsc{SpreadsheetLLM} in accurately understanding complex spreadsheet layouts and structures.
\vspace{-2mm}
\item In order to extend the horizontal capabilities of \textsc{SpreadsheetLLM} to a wide range of downstream tasks, we propose CoS and verify it on Spreadsheet QA, highlighting its potential for intelligent user interaction.
\end{itemize}

\vspace{-2mm}
\section{Related Work}

\label{sec:related}
\paragraph{Spreadsheet Representation}
Spreadsheet representation involves converting the spreadsheets into specific representations for different models. There are various methods for spreadsheet (table) representation. \cite{dong2019semantic,dong2019tablesense} enhance Mask-RCNN to leverage spatial and visual information in spreadsheets, and \cite{deng2024tables} explores the usage of LLMs to evaluate image tables, but it doesn't work well for spreadsheet images as input to VLMs~\cite{xia2024vision}. To capture sequential semantics in rows and columns, LSTMs are further adopted \cite{nishida2017understanding, gol2019tabular} in row\&column directions. Pre-trained LMs~\cite{dong2022tablesurvey} are then proposed to understand spreadsheets~\cite{wang2021tuta}. Recent studies \cite{zhang2023tableLlama,li2023table,sui2023gpt4table} have explored the efficacy of using Markdown, HTML, JSON, and DFLoader for table representation, and these methods are comprehensively summarized in a recent tutorial of~\cite{dong2024large}. However, traditional table representations are not well suited to spreadsheets due to their assumption of single table input and absence of explicit cell addresses, as experiments show in Appendix~\ref{different serialization methods}.

\vspace{-1mm}
\paragraph{Spreadsheet Understanding}
While most table LLMs are restricted to single table settings, spreadsheets with multiple tables typically exceed token limits. Moreover, the diversity in multi-table layout and structure significantly confounds the problem. Spreadsheet table detection~\cite{dong2019tablesense,christodoulakis2020pytheas,doush2010detecting,vitagliano2022mondrian} aims at identifying all tables on a given sheet and determining their respective ranges. As a fundamental task for spreadsheet understanding, this task triggers hundreds of millions of daily average usage in commercial spreadsheet tools~\cite{zhang2024tablellm}, and the accuracy still urges improvements due to the flexibility and complexity of spreadsheets.

\vspace{-2mm}
\paragraph{Spreadsheet Downstream Tasks}
Spreadsheet understanding is enabling for a series of spreadsheet tasks, such as table question answering analysis~\cite{he2024text2analysis,cheng2021hitab,cheng2022binding,jiang2022omnitab,liu2022tapextablepretraininglearning}, table extraction~\cite{chen2013automatic,chen2014integrating,li2024auto}, formula or code generation~\cite{chen2021spreadsheetcoder,cheng2021fortap,joshi2024flame,chen2024auto,li2023SheetCopilot}, error detection~\cite{wang2019uni,dou2016cacheck}, etc. In this paper, we choose spreadsheet QA, one of the most demanded spreadsheet analysis tasks. It is an extension of the Table QA task in spreadsheet data, with the additional complexity of detecting and matching multiple tables within a spreadsheet. 

\vspace{-2mm}
\paragraph{LLMs' Token Efficiency} Related work suggests that the performance of LLMs degrades significantly with long contexts~\cite{liu2024lost, xu2023retrieval}. Efforts to improve model performance and reduce costs have led to the development of compression techniques for long prompts. Some researchers employ information-theory metrics to filter out redundant information~\cite{li2023unlocking, jiang2023llmlingua}. Additionally, specialized models have been proposed to optimize prompt compression~\cite{pan2024llmlingua}. However, these strategies primarily address natural language prompts and may not suit tabular data, potentially leading to considerable structure and data information loss. DBCopilot~\cite{wang2023dbcopilot} enables text-to-SQL conversion on large databases through schema routing. However, due to LLMs' insufficient ability to understand inherent multi-table layouts and complex table structures that cannot execute queries similar to SQL, schema routing is impractical, restricting the broader application of cutting-edge tabular works~\cite{cheng2022binding,li2023table,sui2024table} on spreadsheet data.

\section{Method}
\label{section 4}
We propose a novel spreadsheet encoding framework in a Markdown-like style as text. To achieve a more compact and efficient representation, we introduce three independent yet combinable modules: structural-anchor-based extraction, inverted-index translation, and data-format-aware aggregation, which enable efficient data compression and enhance performance on downstream tasks.

\subsection{Vanilla Spreadsheet Encoding with Cell Value, Address, and Format}
\label{markdown}
Due to the absence of standardized practices in spreadsheet encoding for LLMs, we first explore traditional spreadsheet encoding methods. Appendix \ref{different serialization methods} presents a comparison of different mainstream tabular data encoding methods, including HTML, XML, and Markdown. Based on the encoding length and performance on spreadsheet understanding tasks, we use a Markdown-like style representation:
\vspace{-2mm}
\begin{align}
\mathcal{S} = \{Cell_{i,j}\}_{i\in m,j \in n}, \tag{1}
\end{align}
\vspace{-8mm}
\begin{align}
\mathcal{T} & = \mathrm{markdown} \left\{ \mathrm{encode}  \left( Cell_{i,j} \right) \right \} \nonumber\\  :&= ``|Address_{i,j},V\!alue_{i,j}, F\!ormat|...\setminus \! n", \tag{2}
\end{align}
where $\mathcal{S} \in \mathbb{R}^{m,n} $ denotes the spreadsheet, $\mathcal{T} \in \mathbb{R}^{1}$ denotes the text representation of a cell, and $i$, $j$, $m$, $n$ respectively represent the row and column index of the cell and the row and column range of $\mathcal{S}$. We also explored the inclusion of cell format information (such as background color, bold font, borders, etc.) into each cell's representation. However, these experiments demonstrated that such detailed encoding adversely affects model performance due to rapid token limit exceedance and LLMs' inadequate capability to process format information effectively, as detailed in Appendix \ref{struggle}. We plan to further explore this in future research, focusing on enhancing the model's ability to understand and utilize format and structural cues.

\subsection{Structural-anchor-based Extraction}
Large spreadsheets often feature numerous homogeneous rows or columns, which minimally contribute to the understanding of their layout and structure, as depicted in Figure~\ref{fig:method} (a). To effectively compress spreadsheets while preserving vital layout and structural information, we propose a novel heuristic-based method, detailed further in Appendix \ref{heurisrics model for structural-anchor}. This method identifies heterogeneous rows and columns at the edges of table boundaries—termed structural anchors:

\begin{align}
\small
\mathcal{A} = \left \{ r_p,c_q \right \}_{p \in m, q \in n}, \tag{3}
\end{align}
where {\small $r_p=\left\{ Cell_{i,j}\right\}_{i=p,j \in n}$} and {\small $c_q=\left\{ Cell_{i,j}\right\}_{i \in m,j=q}$}.
Using these anchor points, our method discards rows and columns that are located more than $k$ cells away from any anchor point, because they rarely serve as table boundaries. The parameter $k$ serves as a threshold to control the scope of neighborhood retention, effectively eliminating areas predominantly filled with homogeneous data that do not contribute to an understanding of the spreadsheet's layout and structure. We explored the effects of different $k$ values in an ablation study, as detailed in Appendix \ref{ablation experiment results on threshold}.

The extracted rows and columns can be expressed as:
\begin{align}
\small
\mathcal{A}_{+} = \left \{ r_{p_{+}},c_{q_{+}} \right \}_{p_{+} \in m, q_{+} \in n}, \tag{4}
\end{align}
where the extracted "skeletons" are defined as:  $r_{p_{+}}=\left\{ Cell_{i,j}\right\}_{|i-p| \leq k,j \in n}$ and $c_{q_{+}}=\left\{ Cell_{i,j}\right\}_{i \in m,|j-q| \leq k}$. 
Then we obtain the extracted compact spreadsheet:
\begin{align}
\small
\mathcal{S}_e =  \mathrm{extract} (\mathcal{S}) = \mathrm{address\_map}(r_{p_{+}} \! \cap c_{q_{+}}). \tag{5}
\end{align}
Based on the compressed spreadsheet $\mathcal{S}_e$, we can obtain extremely shorter text representation $\mathcal{T}_e$.
Furthermore, after extraction, we perform a coordinate re-mapping to ensure continuity in cell coordinates, preserving the integrity of data relationships within the compressed spreadsheet. This re-mapping is critical for maintaining the accuracy of prediction results, ensuring that analyses remain consistent even after compression. This method filters out 75\% spreadsheet content but preserves 97\%  rows and columns at the edges of table boundaries. 

\subsection{Inverted-index Translation}

Spreadsheets often contain numerous empty rows, columns, and scattered cells. The standard encoding method, as detailed in Section \ref{markdown}, employs a grid-based method that pairs cell addresses with their contents. This approach necessitates recording empty cells to maintain the spreadsheet's two-dimensional structure, which significantly increases token consumption. Furthermore, cells with identical values are encoded repeatedly, exacerbating token usage.

To address these inefficiencies, we propose a two-stage Inverted-index-based Translation method. The first stage involves converting the traditional matrix-style encoding into a dictionary format, where cell values serve as keys indexing the addresses. In the second stage, cells sharing the same value are merged, with empty cells excluded and cell addresses noted as ranges. This method effectively reduces the number of required tokens by eliminating redundancies and simplifying the representation of repeated and empty cells. The translation process is represented mathematically as follows:
\vspace{-2mm}
\begin{align}
\mathcal{T}_t & = \mathrm{invert} (\mathcal{T}) \nonumber\\
:&= \text{\small $\left \{V\!alue: Address\; or\; Address\_Region, ...\right \}$}. \tag{6}
\end{align}

Inverted-index Translation is a \textbf{lossless compression} method general for all spreadsheet understanding tasks, and it remarkably increases \textsc{SheetCompressor}'s compression ratio from 4.41 to 14.91. More details can be found in Table \ref{Compress Ratio results1}.

\subsection{Data-format-aware Aggregation}
In spreadsheets, adjacent cells typically share the same data format. As shown in Figure \ref{fig:method} (3), column C records the sell-in billed revenue for different products. Nonetheless, the concrete numerical values are not essential for understanding the structure and semantics of the spreadsheet (although there might be a loss of fine-trained details of exact quantities, e.g., "18,476" and "18,674", this does not impact our comprehension that this column represents revenue). In contrast, the data type is critical for understanding spreadsheets. On one hand, data types represent fundamental semantic properties, such as "time" or "phone number". It motivates us to implement rules to match the value of the cell to different data types. On the other hand, in contrast to detailed numerical values, identical data types may be compressed through clustering, thereby reducing the number of tokens.

In this section, we introduce Data-format-aware Aggregation for further compression and information integration. Specifically, we employ Number Format String (NFS), which is a built-in cell attribute in spreadsheets. NFSs can be extracted by default using tools like ClosedXML or OpenPyXL, used to describe the format of cell data as a string. For instance, the NFS for "2024.2.14" is "yyyy-mm-dd", indicating a specific \textbf{date} format. However, spreadsheet users do not always explicitly add NFSs to cells, so NFSs are sometimes absent. As a complement, we propose a rule-based recognizer to map a cell value to a specific predefined data type: {Year, Integer, Float, Percentage, Scientific notation, Date, Time, Currency, Email, and Others}. The first nine types broadly cover approximately 55\% of the cells in our dataset derived from real-world corpora. Finally, based on the NFSs and data type, the aggregator aggregates the cells by Algorithm~\ref{alg:aggregate_indentical_areas}.
This process can be represented as follows:
\vspace{-1mm}
\begin{align}
\small
N\!F\!S\!s = \mathrm{nfs}(\left \{Cell_{i,j} \right \}_{i \in m, j \in n}), \tag{7}
\end{align}
\vspace{-6mm}
\begin{align}
\small
\mathcal{T}_a = \mathrm{aggregator}(\left \{Cell_{i,j} \right \}_{i \in m, j \in n}, N\!F\!S\!s, R), \tag{8}
\end{align}
where $R$ denotes the predefined rules as detailed above. In this way, we further reduce the number of tokens. The compression ratio of the data regions also increases from 14.91 to 24.79. More detailed compression effects of different modules are displayed in Table \ref{Compress Ratio results1}.

\subsection{Chain of Spreadsheet}
\label{cos}
To extend the applicability of \textsc{SpreadsheetLLM} to a broader range of downstream tasks, we introduce the Chain of Spreadsheet (CoS), which unfolds two stages:
\vspace{-2mm}
\paragraph{Table Identification and Boundary Detection} Initially, the compressed spreadsheet and the specific task query are input into the LLM. Leveraging the advances in spreadsheet table detection, the model identifies the table that is relevant to the query and determines the precise boundaries of the relevant content. This step ensures that only pertinent data is considered in the subsequent analysis, optimizing the processing efficiency and focus.
\vspace{-2mm}
\paragraph{Response Generation} The query and the identified table section are re-inputted detected tables into the LLM. The model then processes this information to generate an accurate response to the query. 

Through the CoS, \textsc{SpreadsheetLLM} effectively handles complex spreadsheets by breaking down the process into manageable parts, thus enabling precise and context-aware responses. In this paper, we validate the effect of the Spreadsheet QA task, which is detailed in Section \ref{spreadsheetqa}.

\section{Experiments}
\label{others}
In our experimental evaluation, we first verified the effectiveness of our method in spreadsheet understanding. For this purpose, we chose the classic and foundational task of spreadsheet table detection \cite{dong2019tablesense}. This task serves as a critical benchmark for assessing the framework's ability to accurately identify and interpret table structures within spreadsheets. Building upon this foundational understanding, we further explored the applicability of our method to downstream applications by selecting the representative task of spreadsheet QA. This allows us to test the model's capability to not only detect but also comprehend and respond to user queries based on the data and structure identified in the spreadsheets.

\vspace{-2mm}
\subsection{Spreadsheet Table Detection}

\subsubsection{Dataset}
\label{dataset}

We used the dataset introduced by~\cite{dong2019tablesense}, a benchmark dataset of real-world spreadsheets with annotated table boundaries. Due to the complexity and ambiguity of precise address labeling (the Fleiss Kappa on the test set is 0.830), we further implemented the quality improvement pipeline on the test set by five human professionals, as detailed n in Appendix \ref{Test dataset quality improvement pipeline}. To this end, we obtained a highly validated test set containing 188 spreadsheets. Based on the token usage of the vanilla encoding method, we divided the test set into four categories: Small, Medium, Large, and Huge, with a partition of 64:32:70:22. More details are shown in Appendix \ref{test dataset partition}. We adopted the Error-of-Boundary 0 (EoB-0) metric for evaluation on 188 spreadsheets with 311 tables. EoB-0 requires \textbf{exact match} of the top, left, bottom, and right boundaries.  

\vspace{-1mm}
\subsubsection{Experiment Setup}

\paragraph{Baseline \& Evaluation Metrics}

To evaluate the performance of \textsc{SpreadsheetLLM}, we chose TableSense-CNN~\cite{dong2019tablesense} as the baseline due to its previously demonstrated effectiveness in the spreadsheet table detection task. We employed the micro F1 Score as the primary metric to evaluate and compare the performance of different models, as it balances precision and recall, providing a holistic view of model accuracy.

\vspace{-2mm}
\paragraph{Model Selection} 
The experiments included both closed-source and open-source models. From the closed-source spectrum, we selected two versions of OpenAI's models: GPT4 and GPT3.5, which are known for their advanced language understanding capabilities. On the open-source side, we chose Llama2, Llama3, Phi3, and Mistral-v2. The specific configurations are detailed in Appendix \ref{experiment setups and parameters}. 
\subsection{Spreadsheet QA}
\label{spreadsheetqa}
\subsubsection{Dataset}

Existing datasets for the Table QA task focus solely on single-table scenarios, leaving a notable gap in performance evaluation for spreadsheets that contain multiple tables. To bridge this gap, we developed a new Spreadsheet QA dataset tailored to the complexities of multi-table environments. We sampled 64 spreadsheets from our larger collection and crafted 4-6 questions per spreadsheet, targeting fundamental operations such as searching, comparison, and basic arithmetic. We deliberately excluded questions involving composite operations to maintain clarity and focus in testing specific skills. Each question was paired with an answer, formatted either as a specific cell address or a formula that includes cell addresses, facilitating direct and unambiguous evaluations of the model's ability to navigate and interpret spreadsheet data. This approach resulted in a comprehensive test dataset comprising 307 items, each a tuple of $(Q, A, S)$, which is detailed in Appendix \ref{Spreadsheet QA dataset}.

\begin{table*}[htbp]
\begin{center}

  \centering
  \scalebox{0.73}{
  \begin{tabular}{lc|cccccccc}
    \toprule
     & Metric & No Modules & Module 1 & Module 2 & Module 3 & Module 1\&2 & Module 1\&3 & Module 2\&3 & Module 1\&2\&3 \\

    \midrule
        & Total Tokens & 1,548,577 & 350,946 & 580,912 & 213,890 & 103,880 & 96,365 & 211,445 & 62,469 \\
        & Compression Ratio & 1.00 & 4.41 & 2.67 & 7.24 & 14.91 & 16.07 & 7.32 & 24.79 \\

    \bottomrule
  \end{tabular}
  }
  \caption{Average Compression Ratio on test datasets. Results of the train \& valid set are shown in Appendix \ref{compression ratio results appendix}.}
  \label{Compress Ratio results1}
\end{center}
\end{table*}

\subsubsection{Experiment Setup}

\paragraph{Baseline \& Evaluation Metrics} Given that LLMs have not yet been systematically applied to Spreadsheet QA tasks, we have selected \textsc{TaPEx} and Binder~\cite{liu2022tapextablepretraininglearning, cheng2022binding}, which are established baselines in the Table QA domain, for comparative evaluation. Since \textsc{TaPEx} and Binder are designed primarily for single-table data, we adapted them for our multi-table context. Initially, our fine-tuned model identifies table regions relevant to each question. These regions are then formatted and fed into the baseline models. In cases where the input exceeds the token limitations of the baseline models, truncation is employed. The accuracy of the answers is assessed based on the correctness of the cell addresses and cell combinations/calculations provided in the answers. 

\vspace{-2mm}
\paragraph{Model Selection} Our experiments were conducted using the GPT4 model, leveraging its advanced capabilities in language understanding and reasoning. Details on parameters and configurations used are documented in Appendix \ref{experiment setups and parameters}.

\subsubsection{Experiment Procedure}
In this section, we employed the model fine-tuned on the spreadsheet table detection task to conduct QA experiments. In the QA experiments, we utilized the fine-tuned model on the table boundary detection task for spreadsheet QA as an ablation to study the generalization capability of the fine-
tuned boundary detection model. CoS supports multi-step reasoning during the response generation process. In the QA scenarios, the whole steps include spreadsheet table detection, table structure understanding, table splitting, and sub-table QA. 
In cases where tables were exceptionally large and defy effective compression, we utilized a table-splitting algorithm designed to recognize headers and perform strategic concatenation, ensuring that each segment of the split table retains as much contextual integrity as possible. The specifics of this algorithm are detailed in Appendix \ref{Table Split QA Algorithm}. We categorically classify the model's responses as either correct or incorrect, and we select accuracy as the evaluation metric.

\section{Results}
\subsection{Compression Ratio} 
The effectiveness of our encoding process in reducing the size of spreadsheet data is quantitatively assessed using the compression ratio, which is defined by the formula:
\begin{align}
\small
r = n / n', \tag{9}
\end{align}
Our encoding methodology has significantly optimized token usage within spreadsheets. In our test set, it achieved an impressive 25$\times$ compression ratio, substantially reducing the computational load for processing large datasets. The specific compression ratios achieved by various module combinations within \textsc{SheetCompressor} are detailed in Table \ref{Compress Ratio results1}. These results highlight the efficacy of our approach across different configurations, demonstrating its robustness and adaptability in handling diverse spreadsheet structures.

\subsection{Spreadsheet Table Detection}
\begin{table}[t]
\centering
  \begin{adjustbox}{scale=0.75,center}
  \begin{tabular}{lccccc}
    \toprule
    Model \& Method & Small & Medium & Large & Huge & All \\
    \midrule
    \multicolumn{6}{l}{\large \textbf{ICL}} \\ 
    \midrule
    Mistral-v2 & 0.071 & 0.013 & 0.029 & 0.017 & 0.036 \\
    GPT4 & 0.318 & 0.292 & 0.090 & 0.000 & 0.154 \\
    GPT4-compress & 0.480 & 0.454 & 0.373 & 0.330 & 0.410 \\
    \midrule
    \multicolumn{6}{l}{\large \textbf{Fine-tune}} \\
    \midrule
    Llama3 & 0.715 & 0.765 & 0.290 & 0.000 & 0.471 \\
    Llama2 & 0.557 & 0.378 & 0.107 & 0.000 & 0.280 \\
    Phi3 & 0.604 & 0.481 & 0.201 & 0.130 & 0.330 \\
    Mistral-v2 & 0.700 & 0.784 & 0.472 & 0.123 & 0.542 \\
    GPT4 & 0.779 & 0.707 & 0.288 & 0.000 & 0.520 \\
    \midrule
    Llama3-compress & 0.825 & 0.768 & 0.664 & 0.617 & 0.719 \\
    Llama2-compress & 0.710 & 0.722 & 0.633 & 0.578 & 0.660 \\
    Phi3-compress & 0.800 & 0.673 & 0.624 & 0.675 & 0.689 \\
    Mistral-v2-compress & 0.778 & 0.729 & 0.686 & 0.744 & 0.726 \\
    GPT3.5-compress & 0.795 & 0.649 & 0.600 & 0.680 & 0.680 \\
    GPT4-compress & 0.810 & \textbf{0.832} & 0.718 & 0.690 & 0.759 \\
    \quad -w/o Aggregation & \textbf{0.864} & 0.816 & \textbf{0.739} & \textbf{0.753} & \textbf{0.789} \\

    \midrule
    TableSense-CNN & 0.785 & 0.788 & 0.567 & 0.561 & 0.666 \\

    \bottomrule
  \end{tabular}
  \end{adjustbox}
  \caption{Results of various Model \& Method configurations on spreadsheet table detection. Our encoding method achieved \textbf{SOTA} on the GPT4 model.}
  \label{tab:boundary_detection}
\end{table}

\subsubsection{Main Results}
Table \ref{tab:boundary_detection} illustrates the performance differences among various models and methods on spreadsheet table detection task, and the detailed case study can refer to Appendix \ref{case_all}.

\textbf{1) Enhanced Performance with various LLMs:} The fine-tuned GPT4 model achieved the F1 score of approximately 76\% across all datasets, while our encoding method without aggregation achieved the F1 score of approximately 79\% across all datasets. This marked a 27\% improvement over the same model fine-tuned on original data, a 13\% increase over TableSense-CNN, and established a new \textbf{SOTA}. The entire encoding method slightly reduced the F1 score within a tolerable range, but achieved good compression results, as shown in Table \ref{Compress Ratio results1}. We also evaluated our encoding method on a series of open-source models. Notably, Llama3 and Mistral-v2 achieved an F1 score of approximately 72\%, just 6 percentage points below the SOTA. The improvements due to our compression method were substantial, with increases of 25\% for Llama3, 36\% for Phi3, 38\% for Llama2, and 18\% for Mistral-v2. These results underscored the significant enhancement performance attributable to our encoding method.

\textbf{2) Benefits for Larger Spreadsheets:} Our compression method significantly boosted performance on larger spreadsheets, where the challenges were most pronounced due to model token limits. The improvements in F1 scores were particularly notable on huge spreadsheets (75\% over GPT4, 19\% over TableSense-CNN), large spreadsheets (45\% and 17\%), medium (13\% and 5\%), and small (8\%) spreadsheets. This demonstrated our method's effectiveness in enabling LLMs to process a broader range of spreadsheet sizes efficiently.

\textbf{3) Improvements in In-Context Learning:} Compact encoding also significantly enhanced ICL capabilities. For instance, the performance of GPT4 on all data improved by nearly 26\%, demonstrating the method’s effectiveness beyond fine-tuned models to include ICL scenarios as well. More ICL results are shown in Appendix \ref{open_icl}.

\textbf{4) Significant Cost Reduction:} 
Our cost was almost directly proportional to the input tokens, because the output table regions are short, which can be neglected. Based on the prices of the GPT4 and GPT3.5-turbo models~\footnote{https://azure.microsoft.com/en-us/pricing/details/cognitive-services/openai-service/} in ICL, we reduced 96\% cost in our test set. Detailed calculations are presented in Appendix \ref{Cost calculation}.

\subsubsection{Ablation Experiment Results}

\begin{table}[t]
\begin{center}

  \centering
  \scalebox{0.73}{
  \begin{tabular}{l|ccccc}
    \toprule
    Model & \multicolumn{1}{c}{Small} & \multicolumn{1}{c}{Medium} & \multicolumn{1}{c}{Large} & \multicolumn{1}{c}{Huge} & \multicolumn{1}{c}{All} \\
    \midrule
    GPT4 & 0.779 & 0.700 & 0.288 & 0.000 & 0.520 \\
    GPT4-compress & 0.810 & 0.832 & 0.718 & 0.690 & 0.759 \\
    \quad-w/o Extraction & 0.805 & 0.772 & 0.618 & 0.321 & 0.655 \\
    \quad-w/o Translation & 0.785 & 0.804 & 0.729 & 0.636 & 0.743 \\
    \quad-w/o Aggregation & 0.864 & 0.816 & 0.739 & 0.753 & 0.789 \\
    \bottomrule
  \end{tabular}
  }
  \caption{Ablation studies on spreadsheet table detection.}
  \label{Spreadsheet table detection ablation results1}
\end{center}
\end{table}

Table \ref{Spreadsheet table detection ablation results1} presents the results of ablation experiments for different modules. 
The removal of the extraction module led to significantly lower F1 scores, underscoring its critical role in capturing and retaining key structural information. As highlighted in Table \ref{Compress Ratio results1}, this module also achieved the most significant token reduction, confirming its effectiveness. After removing the aggregation module, the F1 score slightly increased. This observation might be attributed to the NFS being more abstract than straightforward numerical representations, which can challenge an LLM's ability to interpret them effectively. Despite this, the NFS method offered a significantly high compression rate, enhancing its potential for practical applications and cost control. 

\subsection{Spreadsheet QA}

\begin{table}[t]
  \centering
  \footnotesize

  \begin{tabular}{lc}
    \toprule
    Model  & Accuracy \\
    \midrule
    \textsc{TaPEx} & 0.378 \\
    Binder & 0.622 \\
    GPT4  & 0.466 \\
    
    GPT4-compress-w/o splitting & 0.651 \\
    GPT4-compress-w/o splitting-FT & 0.694 \\
    GPT4-compress & 0.684 \\
    GPT4-compress-FT & \textbf{0.743} \\
   
    \bottomrule
  \end{tabular}
  \caption{The results for Spreadsheet QA show that our method achieved SOTA. "-FT" means fine-tuned model on spreadsheet table detection task and is applied to QA.}
  \label{Spreadsheet_QA_results}
\end{table}

Table \ref{Spreadsheet_QA_results} shows the performance of various models on Spreadsheet QA tasks. We can draw the following conclusions:
\textbf{1) Effectiveness of the CoS Method:} The CoS method with \textsc{SpreadsheetLLM} significantly boosted the accuracy of models, showing a notable increase of 22\% over the baseline GPT4 model. Given the large size of typical spreadsheets, directly inputting entire files often exceeded the token limits of conventional models. Some large tables even exceed the token limitation after compression. The CoS effectively addressed this issue by focusing only on regions relevant to the questions posed, improving accuracy by 3\% and 5\% on ICL and fine-tuning respectively.

In structural-anchor-based extraction, some information loss may occur, which could potentially affect downstream tasks. The CoS method was proposed to address this issue. Based on the overall understanding of the spreadsheet and the table detection, CoS identifies the related table and re-encodes it for downstream tasks without structural-anchor-based extraction. Additionally, the three modules proposed in \textsc{SpreadsheetLLM} are modular and can be freely chosen and used according to different tasks. For instance, the invert-index translation module does not result in any information loss and is almost applicable to all downstream tasks.

In the works of \textsc{TaPEx} and Binder that we referenced, tables are unlike spreadsheets, which can feature a multi-table layout, hundreds of rows (or even more), and formats such as merged cells that complicate the understanding of spatial information. To mitigate these challenges and enhance their capabilities on QA tasks, we have used tables detected in the spreadsheet that were related to the query as inputs. On the other hand, the address information of large tables in spreadsheets is crucial. In \textsc{SpreadsheetLLM}, we have augmented each cell with coordinate information. We speculate that if \textsc{TaPEx}'s pre-training had considered this setting, its performance might have further improved.

\textbf{2) Generalization Capability of the Fine-tuned Model:} The model that has been fine-tuned on the spreadsheet table detection task demonstrated robust generalization capabilities across downstream QA tasks. This fine-tuning on table detection led to an accuracy improvement of 6\% on QA. Moreover, it significantly outperformed the \textsc{TaPEx} and Binder models by 37\% and 12\%, respectively. This substantial margin highlighted that fine-tuning not only prepared the model to better understand the specific data and structural nuances of spreadsheets but also enhanced its overall comprehension abilities.

\vspace{-1mm}
\section{Conclusion}

In this paper, we proposed the \textsc{SpreadsheetLLM}, a novel framework representing a significant advancement in the processing and understanding of spreadsheet data by leveraging the capabilities of LLMs. Through a novel encoding method, \textsc{SheetCompressor}, this framework effectively addresses the challenges posed by the size, diversity, and complexity inherent in spreadsheets. It achieves a substantial reduction in token usage and computational costs, enabling practical applications on large datasets. The fine-tuning of various cutting-edge LLMs further enhances the performance of spreadsheet understanding. Moreover, Chain of Spreadsheet, the framework's extension to spreadsheet downstream tasks illustrates its broad applicability and potential to transform spreadsheet data management and analysis, paving the way for more intelligent and efficient user interactions. 

\clearpage
\section*{Limitations}
While \textsc{SpreadsheetLLM} has markedly advanced how LLMs interpret and utilize spreadsheets, they also illuminate areas ripe for further research and development. Currently, our methods do not yet harness spreadsheet format details such as background color and borders, because they take too many tokens. However, these elements often contain valuable contextual and visual cues that could further refine our understanding and processing of spreadsheet data. Additionally, while \textsc{SheetCompressor} effectively aggregates data regions, it does not currently employ a sophisticated semantic-based compression method for cells containing natural language. For example, categorizing terms like "China," "America," and "France" under a unified label such as "Country" could not only increase the compression ratio but also deepen the semantic understanding of the data by LLMs. Exploring these advanced semantic compression techniques will be a key focus of our ongoing efforts to enhance the capabilities of \textsc{SpreadsheetLLM}.

\section*{Ethics Statement}
All data were collected, analyzed, and reported without any bias or influence from external sources. The privacy and confidentiality of the participants were strictly maintained throughout the research process. No personal identifiers were used in the analysis or reporting of the data to ensure anonymity. At the same time, data standard personnel were paid according to the highest local standard, and their daily working hours were strictly limited to no more than 8 hours to protect their legitimate rights and interests.
We acknowledge the contributions of all individuals and institutions involved in this study and are committed to sharing our findings and methodologies transparently to facilitate further research and knowledge advancement in the field.
\bibliography{custom}

\clearpage
\appendix

\section{GPT4 Struggles to Understand Spreadsheets}
\label{struggle}
\begin{figure}[htbp]
    \centering
    \includegraphics[width=0.5\textwidth]{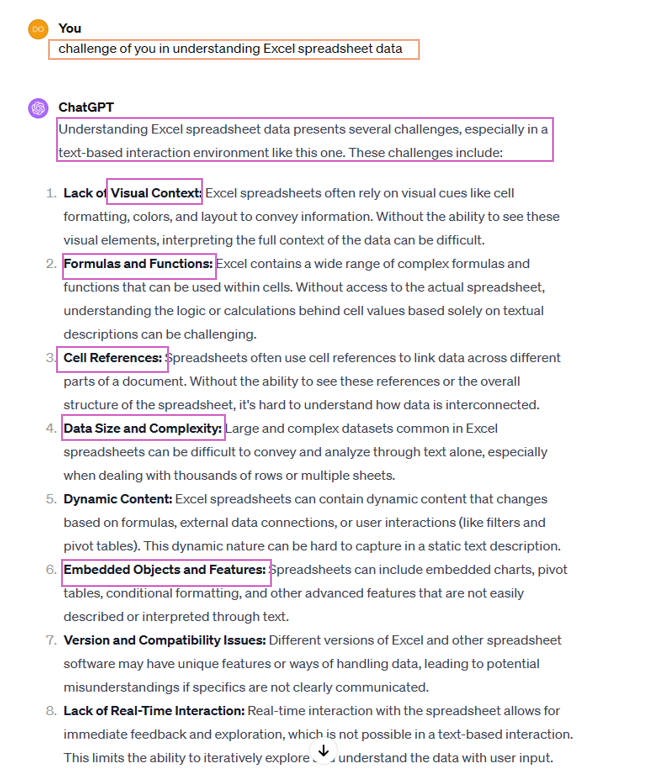}
    \caption{Challenges of GPT4 understanding spreadsheet data.}    
    \label{gpt41}
\end{figure}

\begin{figure}[htbp]
    \centering
    \includegraphics[width=0.5\textwidth]{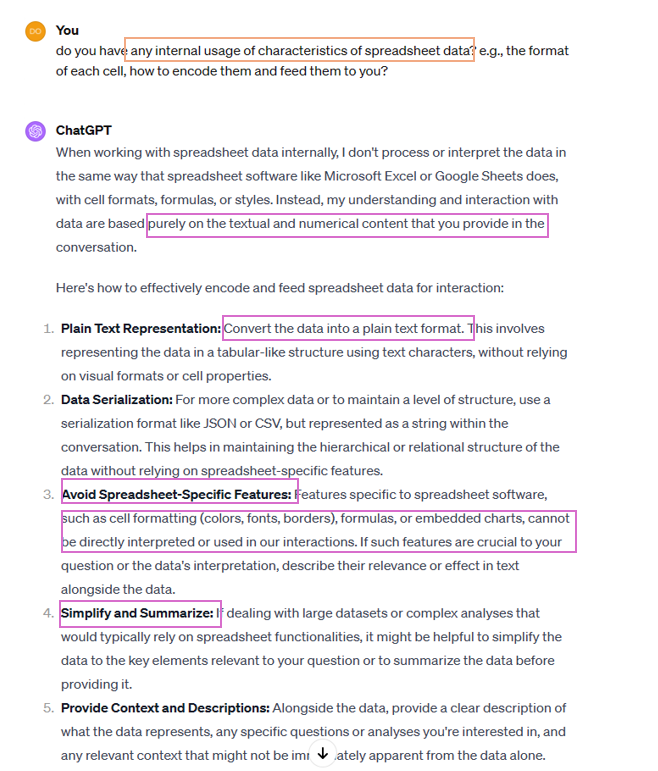}
    \caption{GPT4 encoding methods and techniques for processing spreadsheet data.} 
    \label{gpt42}
\end{figure}

\vspace{-3mm}
The Figure~\ref{gpt41} and Figure~\ref{gpt42} show how GPT4 struggles to understand spreadsheets. We also validated the effect of cell format on the vanilla encoding method on the spreadsheet table detection task. As shown in Table \ref{Spreadsheet table detection ablation fmt results}, the results indicate that in ICL, the inclusion of format marginally improves the model's performance on small datasets but results in poorer performance on larger datasets. For the fine-tuned model, the inclusion of format information leads to a significant reduction in the overall F1 score. This decline is attributed to the introduction of additional tokens, which causes some data to exceed the model's token limits. Additionally, LLMs are not yet adept at understanding format information.

\vspace{-3mm}

\begin{table}[h]
\begin{center}
  \centering
  \scalebox{0.77}{
  \begin{tabular}{l|ccccc}
    \toprule
    Model & \multicolumn{1}{c}{Small} & \multicolumn{1}{c}{Medium} & \multicolumn{1}{c}{Large} & \multicolumn{1}{c}{Huge} & \multicolumn{1}{c}{All} \\
    \midrule
    GPT4-ICL & 0.318 & 0.292 & 0.090 & 0.000 & 0.154 \\
    GPT4-ICL-FMT & 0.429 & 0.000 & 0.000 & 0.000 & 0.204 \\
    GPT4-FT & 0.779 & 0.707 & 0.288 & 0.000 & 0.520 \\
    GPT4-FT-FMT & 0.758 & 0.000 & 0.000 & 0.000 & 0.315 \\
    \bottomrule
  \end{tabular}
  }
\caption{The results of spreadsheet table detection experiment with cell format.}
  \label{Spreadsheet table detection ablation fmt results}
  
\end{center}
\end{table}

\vspace{-4mm}

\section{Traditional Encoding Methods for Spreadsheets}
\label{different serialization methods}
\begin{figure*}[htbp]
    \centering
    \includegraphics[width=1.0\textwidth]{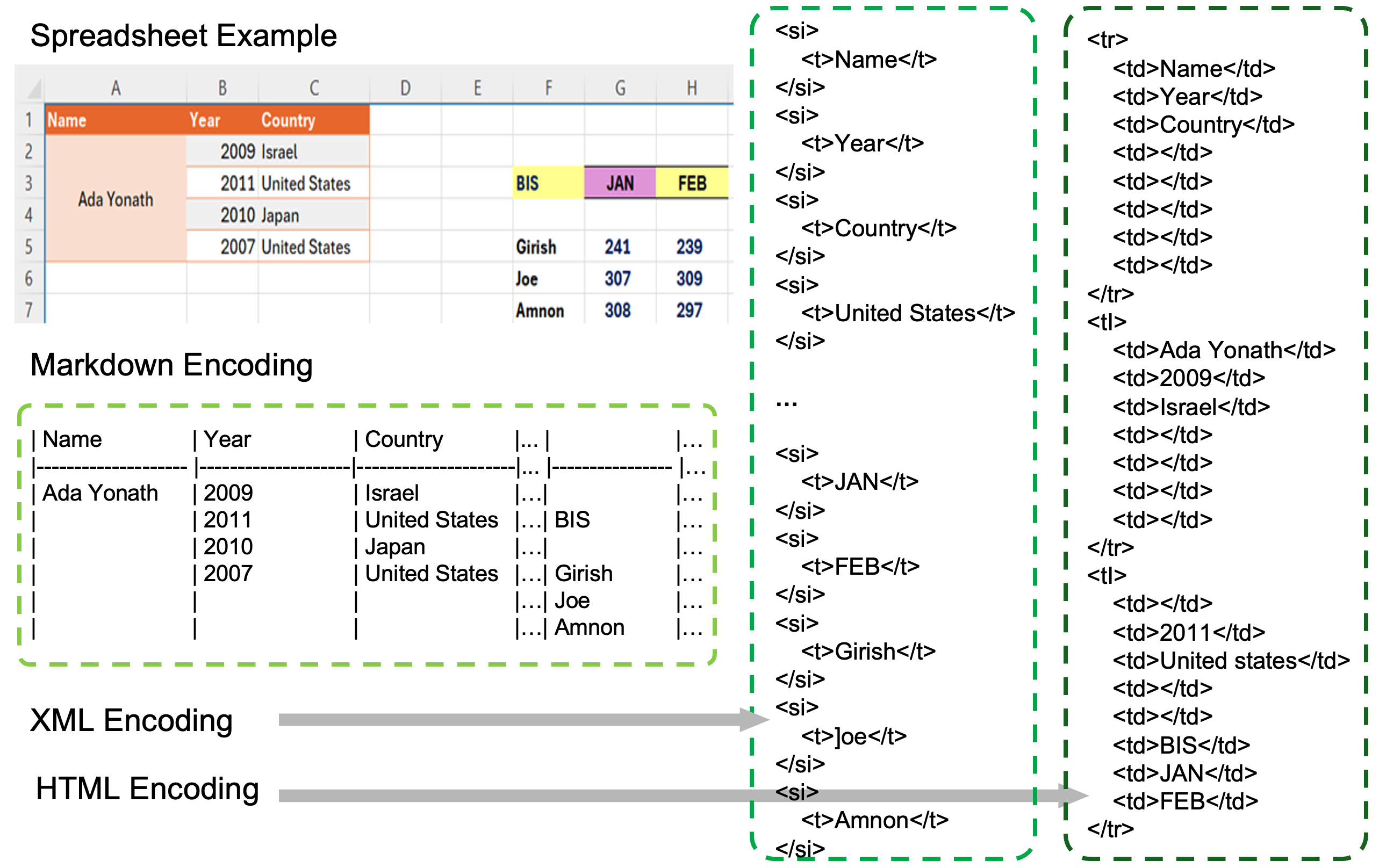}
    \caption{Examples of three traditional spreadsheet encoding methods: Markdown, XML, and HTML. Due to space limitations, we only show the encoding of some cells.}    
    \label{different encoding example}
\end{figure*}

\vspace{-2mm}
In our study, we explored traditional encoding methods—Markdown, XML, and HTML—to represent spreadsheet data. Figure \ref{different encoding example} illustrates the comparative analysis of these methods. XML and HTML encoding, while widely used, tend to result in high token consumption due to the extensive use of repeated label tags necessary for representing the data structure. This approach markedly increases the volume of data processed.

Conversely, the Markdown method, although more token-efficient, has its limitations. One significant drawback is the lack of explicit cell address information, which frequently leads to errors when indexing specific cell locations. Additionally, Markdown's rigid structure rules complicate the accurate representation of merged cells, a common feature in complex spreadsheets that is crucial for preserving the integrity of data relationships.

To quantitatively assess these methods, we conducted ICL experiments using the GPT4 model on spreadsheet detection tasks. The results, detailed in Table \ref{different serialization results}, confirmed that while the Markdown method outperformed XML and HTML in terms of lower token usage, it still fell short in addressing the needs of spreadsheet encoding effectively.

\begin{table}[h]
\begin{center}

  \centering
  \scalebox{0.8}{
  \begin{tabular}{l|ccccc}
    \toprule

        & \multicolumn{1}{c}{Small}    
         & \multicolumn{1}{c}{Medium}    
         & \multicolumn{1}{c}{Large} 
         & \multicolumn{1}{c}{Huge}
         & \multicolumn{1}{c}{All} \\
    \midrule
    HTML & 0.074 &0.016 & 0.003 & 0.000 & 0.031\\
    XML &0.292 &0.102 & 0.066 & 0.000& 0.142\\
    Markdown &0.254 & 0.167 & 0.143 & 0.121 & 0.175\\
    \bottomrule
  \end{tabular}
  }
  \caption{The ICL experiments of different encoding methods of the spreadsheet on GPT4.}
  \label{different serialization results}
\end{center}

\end{table} 

\vspace{-18mm}

\section{Lightweight Heuristics for Structural-anchor Proposal}
\label{heurisrics model for structural-anchor}
Initially, this method enumerates bounding lines by finding discrepancies in neighboring rows and columns based on differences in cell values, merged cells, borders, and fill color. In other words, it enumerates rows and columns with imbalances (text, merge, border, color, font, etc.). Rows and columns without significant discrepancies are usually canonical data rows or columns that contribute trivially to the layout understanding of a spreadsheet. Subsequently, it composes all possible candidate boundaries using any two rows and any two columns as top/bottom/left/right edges. In the third step, heuristics are applied to filter unreasonable boundary candidates by judging the integrity within each candidate boundary. For example, the proportion of numbers and characters in each row and column is used to infer the sparsity in the internal region and four edges of the candidate boundary. The size of the boundary is used to infer if it is too small to be a table, and the presence of header-like rows and columns is also considered. After this step, a small proportion of candidate boundaries are preserved.

\begin{figure}[h]
    \centering
    \includegraphics[width=\linewidth]{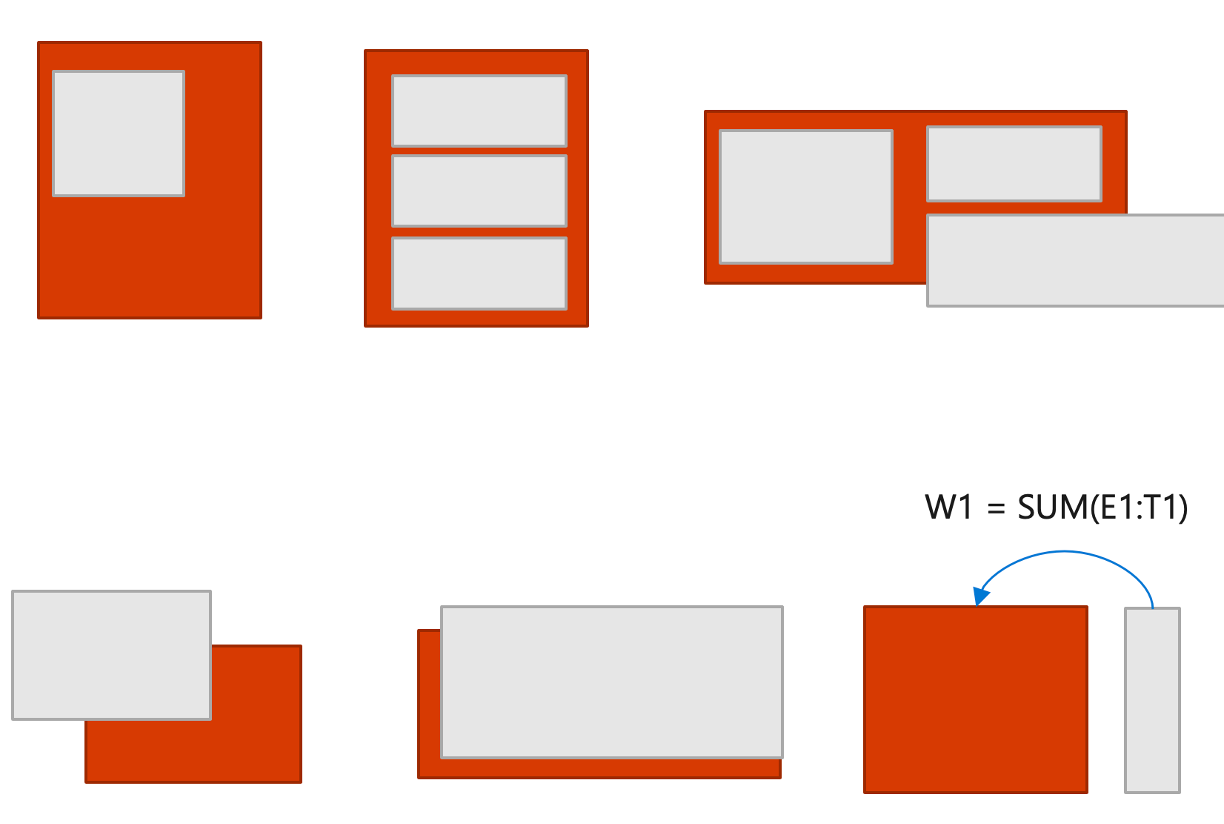} 
    \caption{Common overlapping patterns of candidate boundaries. }
    \label{fig:overlap}
\end{figure}

In the fourth step, overlapping candidate boundaries are enumerated pairwise. Information such as the relative positions of candidate boundaries and the presence of headers is used to determine which candidate boundary more likely represents a table, thereby filtering out some overlapping boundaries. Figure~\ref{fig:overlap} presents common overlapping patterns. For example, for two overlapping candidate boundaries with close top boundaries, heuristics use the proportion of textual cells or format strings like year and date to determine candidate headers. The existence of candidate headers is then used to decide which candidate boundary to filter out. 

Finally, we take the remaining candidate boundaries to derive structural anchors. However, due to the challenge of fine-grained discriminating headers, titles, and notes for heuristics, the candidate boundaries produced by the above heuristics only achieve 46.3\% F1 score in EoB-0 metric and 65.0\% EoB-2 metric in our boundary detection test set. 
Fortunately, including neighboring rows and columns largely increases the coverage of real bounding rows and columns, because headers, titles, and notes usually span few rows and columns. So we propose to not only use the exact bounding rows and columns as structural anchors but also include rows and columns within $k$ rows and columns neighboring the structural anchors to preserve titles, notes, and headers as much as possible. This allows LLMs to further determine the exact boundaries by leveraging their advanced capabilities in semantic understanding and reasoning. When $k$ is set to 4, over 97\% of the border rows and columns in ground truth tables are preserved. This ensures that structure anchors rarely lose critical information about the table skeleton. 

\section{Ablation Experiment Results of Spreadsheet Table Detection}

\subsection{Results on Structure-anchor Threshold}
\label{ablation experiment results on threshold}

Table \ref{Spreadsheet table detection ablation results2} details the ablation study concerning the number of rows and columns retained near candidate boundaries. Optimal results were observed when four rows/columns were preserved, yielding the highest F1 score across all datasets. This outcome is likely due to a balance between preserving essential boundary information and maintaining a feasible compression ratio. Retaining fewer rows/columns might omit critical boundaries, reducing Recall, while preserving more rows/columns diminishes the compression ratio, potentially exceeding the model's token limits.

For smaller data, results indicate a positive correlation between the number of retained rows and the F1 score, suggesting that higher information retention leads to better model performance.

\begin{table}[t]
\begin{center}
  \centering
  \scalebox{0.77}{
  \begin{tabular}{l@{\hspace{10pt}}|ccccc}
    \toprule
    k & \multicolumn{1}{c}{Small}    
      & \multicolumn{1}{c}{Medium}    
      & \multicolumn{1}{c}{Large} 
      & \multicolumn{1}{c}{Huge} 
      & \multicolumn{1}{c}{All} \\
    \midrule
    2 & 0.775& 0.804& 0.686& 0.558& 0.712\\
    4 & 0.810 & 0.832& 0.718& 0.690& 0.759\\
    8 & 0.788& 0.824& 0.773& 0.400 & 0.744\\
    \bottomrule
  \end{tabular}
  }
  \caption{Spreadsheet table detection Ablation on extracted threshold $k$. We present experiment results of three different $k$: 2, 4, and 8 on fine-tuned GPT4.}
  \label{Spreadsheet table detection ablation results2}
\end{center}
\end{table}

\subsection{Results of Spreadsheet Table Detection on ICL}

We conducted experiments on the GPT4, "GPT4-0125-preview" version. As shown in Table \ref{Spreadsheet table detection ablation results3}, the results are consistent with the conclusions we draw from our fine-tuned experiments.
\begin{table}[t]
\begin{center}

  \centering
  \scalebox{0.75}{
  \begin{tabular}{l|ccccc}
    \toprule
    Model & \multicolumn{1}{c}{Small} & \multicolumn{1}{c}{Medium} & \multicolumn{1}{c}{Large} & \multicolumn{1}{c}{Huge} & \multicolumn{1}{c}{All} \\
    \midrule
    GPT4-compress & 0.480 & 0.454 & 0.373 & 0.330 & 0.410 \\
    \quad-w/o Aggregation & 0.386 & 0.271 & 0.215 & 0.267 & 0.280 \\
    \quad-w/o Translation & 0.386 & 0.427 & 0.338 & 0.418 & 0.379 \\
    \quad-w/o Extraction & 0.345 & 0.263 & 0.198 & 0.268 & 0.257 \\
    \bottomrule
  \end{tabular}
  }
  \caption{Ablation experiment results on ICL on spreadsheet table detection. Our compression method also achieved the best F1 score on ICL.}
  \label{Spreadsheet table detection ablation results3}
\end{center}
\end{table}

\section{Spreadsheet Table Detection Test Dataset Quality Improvement Pipeline}
\label{Test dataset quality improvement pipeline}
The quality improvement pipeline on the test set consists of the following steps:
(1) excluding those spreadsheets where at least one cell contains languages beyond English; (2) removing spreadsheets in the test set that lie in the same workbook as at least one spreadsheet in the training set, because spreadsheets in the same workbook, though different, are often similar; (3) annotating all spreadsheets in three types: type 1 means certain for one label; type 2 means multiple labels are reasonable; type 3 means not certain. We employ five well-educated annotators from top universities with majors in computer science to undertake this quality improvement. For each spreadsheet in the test set, we aggregate the annotations from all five annotators and preserve multiple reasonable labeling results for type 2 spreadsheets. 

As a result, we obtained a well-annotated dataset with 167 spreadsheets containing 268 tables for type 1, 21 spreadsheets with 43 tables for type 2, and 10 spreadsheets for type 3. We selected data labeled as type 1 and type 2 as the test set, comprising a total of 188 entries.

\section{Spreadsheet Table Detection Test Dataset Partition}
\label{test dataset partition}
From the spreadsheet raw file, we can extract various features, including cell address, value, format (background color, bold, borders, etc.), and more. We transformed these features into the markdown-like style in Section\ref{markdown}. Then, based on the number of tokens after encoding and the length of the context window of the test model, we divided them into four categories: small (number of tokens less than 4k), medium (4-8k), large (8-32k), and huge (greater than 32k). The following is an example of data in Markdown with format information.

\textbf{Example: Encoding Spreadsheet in Markdown-like Style with Cell Formats}
\begin{mdframed}[backgroundcolor=gray!20, linecolor=black, linewidth=1pt, nobreak=true]
\setlength{\parindent}{0pt} 
Text Input:\\
B2,Table 4: Diesel-driven passenger cars, 2015|C2, |D2, |E2, |F2, |G2, |H2, \\
B3, |C3, |D3, |E3, |F3, |G3, |H3, \\
B4, |C4, |D4, |E4, |F4, |G4, |H4, \\
|B5, |C5,Diesel engine|D5, |E5, |F5,Share of all passenger cars (\%)|G5, |H5,

......

Format Input:

|B2,Font Bold|C2, |D2, |E2, |F2, |G2, |H2, \\
|B3, |C3, |D3, |E3, |F3, |G3, |H3, \\
|B4,Bottom Border,|C4,Bottom Border,|D4,Bottom Border,|E4,Bottom Border,|F4,Bottom Border,|G4,Bottom Border,|H4,Bottom Border,\\
|B5,Top Border,Right Border,Fill Color,Font Bold|C5,Top Border,Bottom Border,Left Border,Fill Color,Font Bold|D5,Top Border,Bottom Border,Fill Color,Font Bold|E5,Top Border,Bottom Border,Right Border,Fill Color,Font Bold|F5,Top Border,Bottom Border,Left Border,Fill Color,Font Bold|G5,Top Border,Bottom Border,Fill Color,Font Bold|H5,Top Border,Bottom Border,Fill Color,Font Bold\\
......
\end{mdframed}

\section{Experiment Setup}
\label{experiment setups and parameters}
Open-source model using Deepspeed for distributed training on a workstation with 8 A100 GPUs by LoRA.

\textbf{Llama2:}meta-Llama/Llama-2-7b-chat-hf;

\textbf{Llama3:}meta-Llama/Meta-Llama-3-8B-Instruct;

\textbf{Mistral-v2:}mistralai/Mistral-v2-7B-Instruct-v0.2;

\textbf{Phi3:}microsoft/Phi-3-mini-128k-instruct;

\paragraph{The parameters of open-source model fine-tuning:}
cutoff len=5800;
learning rate=5e-05;
num train epochs=15.0; 
train batch size=5; 
gradient accumulation steps=8; 
lr scheduler type is cosine; 
max grad norm=1.0; 
warmup steps=0; 
optim is AdamW; 
precision is fp16; 
lora rank=32;
lora alpha=64; 
lora dropout=0.01; 
val size=0.0008;
eval steps=50;
eval batch size=5 

\paragraph{The parameters of GPT4/3.5 model fine-tuning:}
lora\_dim=32,n\_epochs=-1,batch\_size=-1,learning\_rate\_multiplier=1,weight\_decay\_multiplier=1e-05,prompt\_loss\_weight=0,trim\_mode=right.

\paragraph{The parameters of GPT4/3.5 model inference:}
temperature=0, max tokens=300, top p=0.95, frequency penalty=0, presence penalty= 0, stop=None, and the rest are default settings.

\section{Spreadsheet QA Test Dataset}
\label{Spreadsheet QA dataset}
\paragraph{Overall Description}
The dataset of 64 spreadsheets includes 9 single table spreadsheets, 35 double table spreadsheets, 11 spreadsheets containing three tables, and 9 spreadsheets containing four or more tables. Among them, 15 spreadsheets contain fewer than 4k tokens, 20 contain between 4k and 8k, 22 contain between 8k and 32k, and 7 contain more than 32k.

\paragraph{Details of the Dataset Collection }
We selected English-language spreadsheets and invited five well-educated professional annotators to annotate the data.
During selection, spreadsheets containing non-ASCII characters or lacking necessary semantic comprehension information were excluded. 
We ensured that the questions could be answered with relative certainty using the information provided in the tables, minimizing the potential confusion or ambiguity. To further validate the quality of the dataset, we invited two additional annotators to perform cross-verification after the initial question-answer labeling process, ensuring the correctness and rationality of the answers. It shows an answer accuracy of 0.846 in Fleiss Kappa, indicating almost perfect agreement.

\vspace{1mm}
\textbf{Example: Spreadsheet QA Data Item}

\begin{mdframed}[backgroundcolor=gray!20, linecolor=black, linewidth=1pt]
\setlength{\parindent}{0pt} 
QUESTION: "What were the highest temperatures in Washington DC in 1998?"

GROUNDTRUTH:
"X23 AND X24"

PROMPT:
[Instruction + Encoded Spreadsheet]

\end{mdframed}

\vspace{1mm}
\section{Cost calculation}
\label{Cost calculation}
We use the ICL price of GPT4 due to the absence of fine-tuned GPT4's price. We neglect the output sequence since it is much shorter than the input sequence in tasks like spreadsheet boundary detection and QA.
The average cost of processing a spreadsheet in our test set has decreased by 96.0\% for GPT3.5 turbo and GPT4, saving an impressive 96.0\% in costs. The cost reduction similarly applies to all LLMs we used.

\section{Other Experimental Results}

\subsection{Compression Results}
\label{compression ratio results appendix}
Table \ref{Compress Ratio results2} shows the compression ratio of each stage in our method relative to the previous stage.
\begin{table}[b]
\begin{center}

  \centering
  \scalebox{0.595}{
  \begin{tabular}{l|cccc}
    \toprule
     Metric & No Modules & Module 1 & Module 1\&2 & Module 1\&2\&3 \\

    \midrule
         Total Tokens & 1,548,577 & 350,946 & 103,880 & 62,469 \\

         Compression Ratio & 1.00 & 4.41 & 3.38 & 1.66 \\
    \bottomrule
  \end{tabular}
  }
  \caption{Compression Ratio of Data Region. }
  \label{Compress Ratio results2}
\end{center}
\end{table}

Table \ref{Compress_Ratio_results_train_valid} shows the total compression ratio of train and valid datasets.

\begin{table}[t]
\begin{center} 

  \centering
  \scalebox{0.8}{
  \begin{tabular}{l|cc}
    \toprule
     Metric & Before & After  \\
    \midrule
    \multicolumn{3}{l}{\large \textbf{Valid Datasets (200 items)}} \\ 
    \midrule
        Tokens & 1,462,076 & 99,411  \\
        Compression Ratio & 1.00 & 14.71 \\
    \midrule
    \multicolumn{3}{l}{\large \textbf{Train Datasets (7000 items)}} \\ 
    \midrule
        Tokens & 192,879,819 & 11,392,870 \\
        Compress Ratio & 1.00 & 16.93  \\
    \bottomrule
  \end{tabular}
  }
  \caption{Compression performance on train \& valid Datasets. }
  \label{Compress_Ratio_results_train_valid}
\end{center}
\end{table}

\subsection{The ICL results of open-source models on spreadsheet table detection.}
\label{open_icl}
Table \ref{Spreadsheet table detection result5} shows the ICL experiments' F1 score of open-source models on the spreadsheet table detection task. In this experimental setting, the open-source model performs far worse than the closed-source model.
\begin{table}[t]
\begin{center}

  \centering
  \scalebox{0.8}{
  \begin{tabular}{l|ccccc}
    \toprule
    Model & \multicolumn{1}{c}{Small} & \multicolumn{1}{c}{Medium} & \multicolumn{1}{c}{Large} & \multicolumn{1}{c}{Huge} & \multicolumn{1}{c}{All} \\
    \midrule
    Llama3 & 0.042 & 0.028 & 0.020 & 0.018 & 0.027 \\
    Llama2 & 0.062 & 0.023 & 0.038 & 0.027 & 0.041 \\
    Phi3 & 0.037 & 0.040 & 0.041 & 0.000 & 0.034 \\
    Mistral-v2 & 0.071 & 0.013 & 0.029 & 0.017 & 0.036 \\
    \bottomrule
  \end{tabular}
  }
  \caption{The ICL results of open-source models' performance on spreadsheet table detection.}
  \label{Spreadsheet table detection result5}
\end{center}
\end{table}

\subsection{Spreadsheet QA Ablation Experiment}
\label{Spreadsheet_QA_ablation_results}

\begin{table}[t]
  \centering
  \footnotesize
  \begin{tabular}{l|cc}
    \toprule
    Model & Answer & Region \\
    \midrule
    GPT4-compress & 0.743 & 0.974 \\
    \quad -w/o Extraction & 0.716 & 0.892 \\
    \quad -w/o Translation & 0.719 & 0.925 \\
    \quad -w/o Aggregation & 0.726 & 0.928 \\
    \bottomrule
  \end{tabular}
\caption{Each module of \textsc{SheetCompressor} contributes to the positive outcomes on Spreadsheet QA. "Answer" represents the accuracy of answering questions, and "Region" represents the accuracy of predicting relevant regions in CoS.}

  \label{Spreadsheet_QA_ablation_resultstab:sheetcompressor_qa}
  
\end{table}

Table \ref{Spreadsheet_QA_ablation_resultstab:sheetcompressor_qa} assesses the impact of removing individual modules on the QA performance. It details both the overall accuracy and the accuracy of identifying question-related regions during the CoT process. The removal of any module generally leads to a decrease in both metrics, with the most significant drop observed when the extraction module is omitted. This is likely due to the extraction module achieving the lowest compression ratio (see Table \ref{Compress Ratio results1}), suggesting that a longer context may hinder the model’s ability to accurately interpret the data.

\section{Case Study}
\label{case_all}
\subsection{Comparison of results before and after structural-anchor-based extraction}

\begin{figure}[h]
    \centering
    \includegraphics[width=\linewidth]{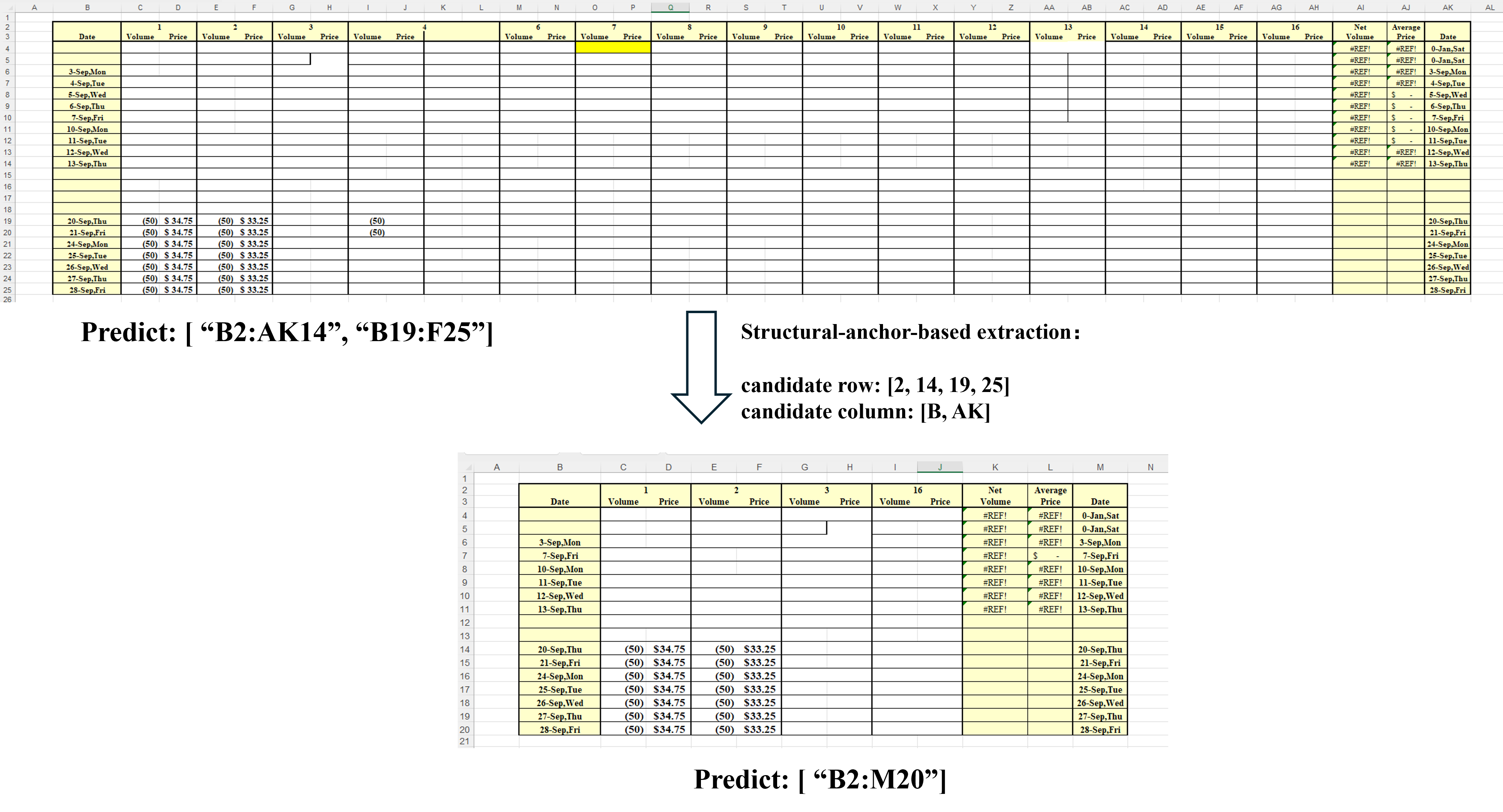} 
    \caption{The results before and after structural-anchor-based extraction. }
    \label{fig:case-study-case1}
\end{figure}

The case described in Figure.~\ref{fig:case-study-case1} illustrates the results of GPT4-FT  before and after structural-anchor-based extraction. Specifically, before structural-anchor-based extraction, most of the content in the spreadsheet is concentrated in the first two rows and the three columns on the left and right, leaving the middle largely empty. This led GPT4 to incorrectly predict the presence of two tables, "B2:AK14" and "B19:F25." However, after applying structural-anchor-based extraction, the spreadsheet's structure becomes more compact, making it easier for GPT4 to correctly predict the table's range as "B2:M20" after coordinating rearrangement.

From this case, we can observe that for spreadsheets with sparse structures and many empty cells, structural-anchor-based extraction not only significantly reduces the number of tokens but also effectively enhances GPT4's ability in table detection.

\subsection{Comparison of results before and after inverted-index translation}

\begin{figure}[h]
    \centering
    \includegraphics[width=\linewidth]{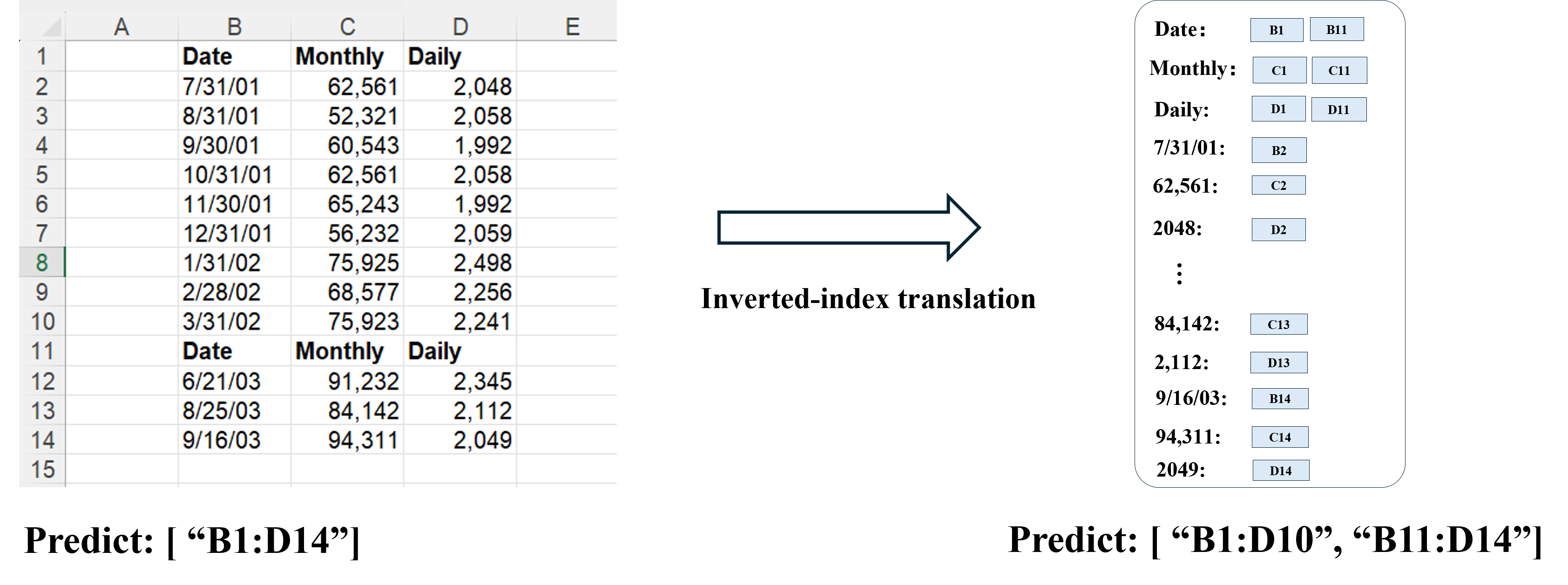} 
    \caption{The results before and after inverted-index translation.}
    \label{fig:case-study-case2}
\end{figure}

The case described in Figure.~\ref{fig:case-study-case2} demonstrates the results of GPT4-FT before and after inverted-index translation. Specifically, before the inverted-index translation, the spreadsheet contained two tables with identical column headers placed closely together, causing GPT4 to mistakenly predict them as one large table, "B1:D14." However, after inverted-index translation, GPT4 was able to aggregate cells with shared values, thereby recognizing semantic relationships between non-adjacent rows and columns. This enabled it to correctly identify the two separate tables in the spreadsheet, "B1:D10" and "B11:D14".

This case indicates that inverted-index translation, by aggregating cells with shared values, not only reduces token redundancy to some extent but also leverages the model's robust understanding of semantic relationships.

\subsection{Comparison of results before and after data-format-aware cell aggregation}
\begin{figure}[h]
    \centering
    \includegraphics[width=\linewidth]{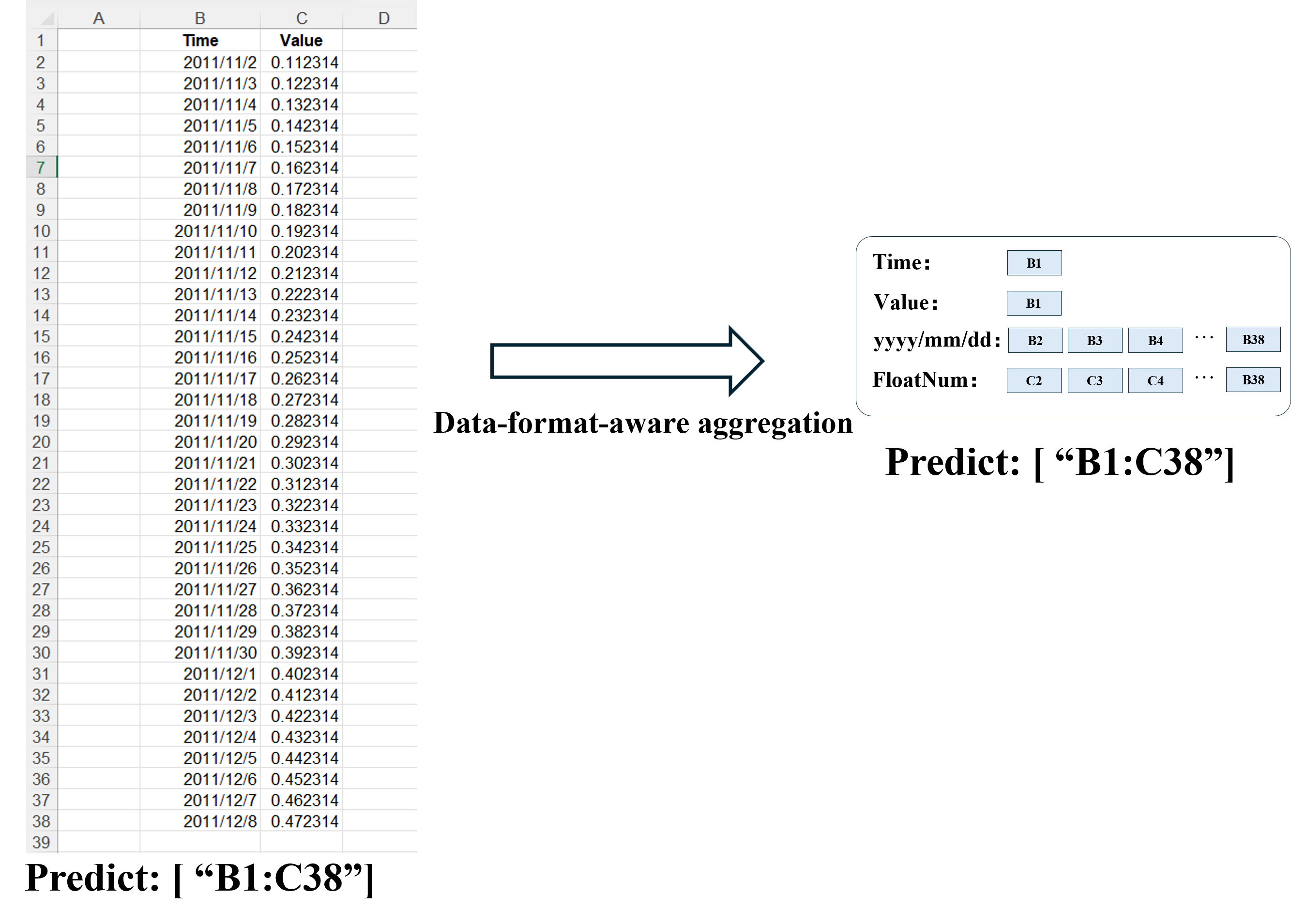} 
    \caption{The results before and after data-aware cell aggregation.}
    \label{fig:case-study-case3}
\end{figure}

The case presented in Figure.~\ref{fig:case-study-case3} showcases the results of GPT4-FT before and after data-aware cell aggregation. Specifically, before data-aware cell aggregation, the spreadsheet contained two columns with values of the same data type, occupying a large number of tokens. The first column increased incrementally by date, while the second column increased incrementally by value. After data-aware cell aggregation, the dates in the first column were replaced with the format string "yyyy/mm/dd" and their addresses were aggregated. Similarly, numerical values were handled with a "FloatNum" format. This method allowed the model to predict the table range correctly as "B1:C38," both before and after processing, indicating that this approach significantly reduces the token count while preserving the semantic information of the spreadsheet data.
\subsection{Comparison of \textsc{SpreadsheetLLM} and TableSense-CNN}
\begin{figure*}[h]
    \centering
    \includegraphics[width=\linewidth]{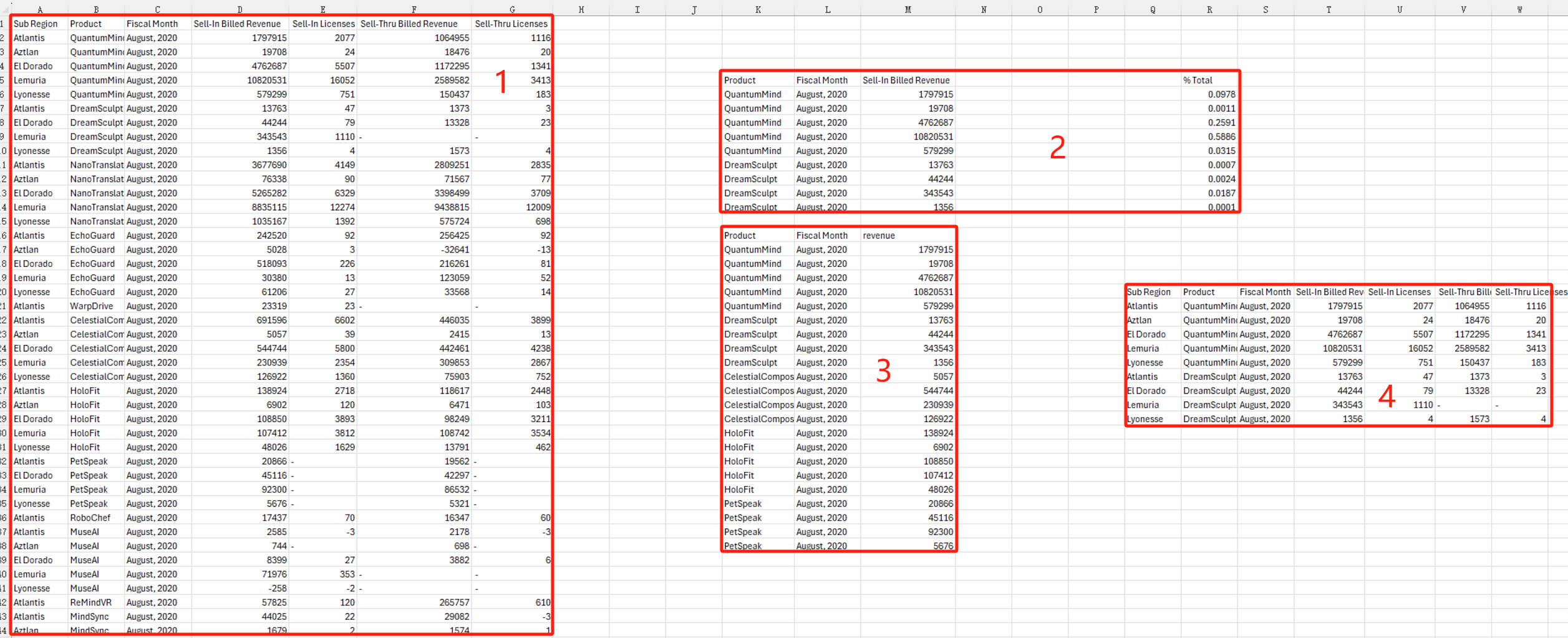} 
    \caption{A challenging case. Traditional spreadsheet understanding methods usually miss the region "R5:R14", but this column has a semantic relationship with the left cells, representing the percentage of those values in left cells.}
    \label{fig_compare}
\end{figure*}

As shown in Figure \ref{fig_compare}, the output of TableSense-CNN is [A1:G44,K5:M14,K16:M38,Q20:W29], while the output of \textsc{SpreadsheetLLM} is [A1:G44,K5:\textbf{R}14,K16:M38,Q20:W29]. \textsc{SpreadsheetLLM} succeeds in adding the region "R5:R14" to the table2. Though it is spatially distant from the table on the left, \textsc{SpreadsheetLLM} can extract the connections from cells' semantic and structural relationship, which demonstrates its powerful reasoning ability.

\section{Prompt Template}
In this section, we present the prompt templates for the Spreadsheet Table Detection and Spreadsheet QA tasks.
\subsection{Vanilla Prompt Template for Spreadsheet Table Detection}
\textbf{A Vanilla Prompt Template for Spreadsheet Table Detection:}
\begin{mdframed}[backgroundcolor=gray!20, linecolor=black, linewidth=1pt, nobreak=true]
\setlength{\parindent}{0pt} 
INSTRUCTION: 

Given an input that is a string denoting data of cells in a spreadsheet. The input spreadsheet includes many pairs, and each pair consists of a cell address and the text in that cell with a ',' in between, like 'A1,Year'. Cells are separated by '|' like 'A1,Year|A2,Profit'. The text can be empty so the cell data is like 'A1, |A2,Profit'. The cells are organized in row-major order. Now you should tell me the range of the table in a format like A2:D5, and the range of the table should only CONTAIN HEADER REGION and the data region, DON'T include the title or comments. Note that there can be more than one table in the string, so you should return all the RANGE, LIKE [{'range': 'A1:F9'}, {'range': 'A12:F18'}]. DON'T ADD OTHER WORDS OR EXPLANATION.

INPUT: 

[Encoded Spreadsheet]
\end{mdframed}

\subsection{Prompt Template for Spreadsheet Table Detection}

\textbf{\quad\textsc{SpreadsheetLLM} Prompt Template for Spreadsheet Table Detection:}

\begin{mdframed}[backgroundcolor=gray!20, linecolor=black, linewidth=1pt, nobreak=true]
\setlength{\parindent}{0pt} 
INSTRUCTION: 

Given an input that is a string denoting data of cells in an Excel spreadsheet. The input spreadsheet contains many tuples, describing the cells with content in the spreadsheet.  
Each tuple consists of two elements separated by a '|': the cell content and the cell address/region, like (Year|A1), ( |A1) or (IntNum|A1:B3). The content in some cells such as '\#,\#\#0'/'d-mmm-yy'/'H:mm:ss',etc., represents the CELL DATA FORMATS of Excel. The content in some cells such as 'IntNum'/'DateData'/'EmailData',etc., represents a category of data with the same format and similar semantics. For example, 'IntNum' represents integer type data, and 'ScientificNum' represents scientific notation type data. 'A1:B3' represents a region in a spreadsheet, from the first row to the third row and from column A to column B. Some cells with empty content in the spreadsheet are not entered. Now you should tell me the range of the table in a format like A2:D5, and the range of the table should only CONTAIN HEADER REGION and the data region. DON'T include the title or comments. 
Note that there can be more than one table in a string, so you should return all the RANGE. 
\end{mdframed}
\begin{mdframed}[backgroundcolor=gray!20, linecolor=black, linewidth=1pt, nobreak=true]
\setlength{\parindent}{0pt} 
DON'T ADD OTHER WORDS OR EXPLANATION.

INPUT: 

[Encoded Spreadsheet] 
\end{mdframed}

\subsection{Prompt Template for Spreadsheet QA}
As detailed in Section \ref{spreadsheetqa}, the CoS method includes two stages, and the prompts for each stage are as follows:

\textbf{Spreadsheet QA Prompt Template:}

\begin{mdframed}[backgroundcolor=gray!20, linecolor=black, linewidth=1pt, nobreak=true]
\setlength{\parindent}{0pt} 
\textbf{Stage 1:} 

INSTRUCTION: 

Given an input that is a string denoting data of cells in a table. The input table contains many tuples, describing the cells with content in the spreadsheet.  Each tuple consists of two elements separated by a '|': the cell content and the cell address/region, like (Year|A1), ( |A1) or (IntNum|A1:B3). The content in some cells such as '\#,\#\#0'/'d-mmm-yy'/'H:mm:ss',etc., represents the CELL DATA FORMATS of Excel. The content in some cells such as 'IntNum'/'DateData'/'EmailData',etc., represents a category of data with the same format and similar semantics. For example, 'IntNum' represents integer type data, and 'ScientificNum' represents scientific notation type data. 'A1:B3' represents a region in a spreadsheet, from the first row to the third row and from column A to column B. Some cells with empty content in the spreadsheet are not entered. How many tables are there in the spreadsheet? Below is a question about one certain table in this spreadsheet. I need you to determine in which table the answer to the following question can be found, and return the RANGE of the ONE table you choose, LIKE [{'range': 'A1:F9'}]. DON'T ADD OTHER WORDS OR EXPLANATION.

INPUT: 

[Encoded Spreadsheet with compression] 
\end{mdframed}
\begin{mdframed}[backgroundcolor=gray!20, linecolor=black, linewidth=1pt, nobreak=true]
\setlength{\parindent}{0pt} 

\textbf{Stage 2:} 

INSTRUCTION: 

Given an input that is a string denoting data of cells in a table and a question about this table. The answer to the question can be found in the table. The input table includes many pairs,

\end{mdframed}
\begin{mdframed}[backgroundcolor=gray!20, linecolor=black, linewidth=1pt, nobreak=true]
\setlength{\parindent}{0pt} 
and each pair consists of a cell address and the text in that cell with a ',' in between, like 'A1,Year'. Cells are separated by '|' like 'A1,Year|A2,Profit'. The text can be empty so the cell data is like 'A1, |A2,Profit'. The cells are organized in row-major order. The answer to the input question is contained in the input table and can be represented by cell address. I need you to find the cell address of the answer in the given table based on the given question description, and return the cell ADDRESS of the answer like '{[B3]}' or '{[SUM(A2:A10)]}'. DON'T ADD ANY OTHER WORDS."

INPUT: 

[Encoded Spreadsheet without compression] 
\end{mdframed}

\clearpage
\section{Algorithm Steps}
\subsection{Identical Cell Aggregation}
\label{Identical Cell Aggregation}
The corresponding algorithm steps are shown in Algorithm~\ref{alg:aggregate_indentical_areas}.

\RestyleAlgo{ruled}
\SetKwComment{Comment}{/* }{ */}
\begin{algorithm}[hbt!]
\caption{Identical Cell Aggregation}
\label{alg:aggregate_indentical_areas}
\SetKwInOut{Input}{Input}
\SetKwInOut{Output}{Output}
\Input{Matrix $nfs$ composed of all cell values in the spreadsheet.  }

\BlankLine

Initialize $m$ and $n$ as the number of matrix $input$ rows and columns.

Initialize the $m \times n$ matrix $visited$ with all values set to $False$.

Initialize $areas$ as an empty list.

Initialize the $FormatDict$ dictionary, the key-value pairs are data values and predefined types respectively.
\BlankLine

\SetKwFunction{FMain}{dfs}
\SetKwProg{Fn}{Function}{:}{}
\Fn{\FMain{r, c, val\_type}}{
    \If{ $visited[r][c]) \lor val\_type != \text{FormatDict}[nfs[r,c]]$}{
        \Return $[r, c, r-1, c-1]$\;
    }
    $visited[r][c] \leftarrow \text{True}$\;
    $bounds \leftarrow [r, c, r, c]$\;
   \ForEach{(tr, tc) around (r, c)}{
        \If{ $\neg visited[tr][tc]) \land val\_type == \text{FormatDict}[nfs[tr,tc]]$}{
            $new\_bounds \leftarrow \text{dfs}(tr, tc, val\_type)$\;
            update bounds to include new\_bounds\;
        }
    }
    \Return bounds\;
}

\BlankLine
\For{$r = 0$ \KwTo $m-1$}{
    \For{$c = 0$ \KwTo $n - 1$}{
        \If{$\neg visited[r][c] $}{
                $val\_type \leftarrow \text{FormatDict}[nfs[r,c]]$\;
                $bounds \leftarrow \text{dfs}(r, c, val\_type)$\;
                    $areas \leftarrow areas + ((bounds[0], bounds[1]), $
                    
                    $ \ (bounds[2], bounds[3]),$ 
                    
                    $ \ val\_type);$

        }
    }
}

\Output{Aggregation matrix $areas$, each cell which is filled with the corresponding datatype after applying custom rules.}

\end{algorithm}

\subsection{Table Split QA Algorithm}
\label{Table Split QA Algorithm}
The corresponding algorithm steps are shown in Algorithm~\ref{alg:qa_large_tables}.

\RestyleAlgo{ruled}
\SetKwComment{Comment}{/* }{ */}
\begin{algorithm}[hbt!]
\caption{Question Answering Process for Large Tables}
\label{alg:qa_large_tables}
\SetKwInOut{Input}{Input}
\SetKwInOut{Output}{Output}

\Input{$question$ composed of strings and two-dimensional matrix $region$}

Initialize $header$ and $answers$ to empty lists

\If{$calculateTokens(region) \leq 4096$} {
    \Return answer\_question(question, region)\;
}
\Else {
    $header \gets predict\_header(region)$\; 
    $body \gets region[length(header) + 1:end]$\;
    
    \For{$i = 0$ \KwTo $length(body)$} {
        $new\_table \gets header + body[i:i+3]$\; 
        $answer \gets answer\_question(question, table)$\;
        $answers.append(answer)$\;
    }
}
\Output{final result $answers$}
\end{algorithm}

\end{document}